\documentclass[10pt,twocolumn,letterpaper]{article}

\usepackage{cvpr}              

\usepackage{graphicx}
\usepackage{svg}
\usepackage{amsmath}
\usepackage{amssymb}
\usepackage{booktabs}
\usepackage[numbers]{natbib}
\usepackage{xcolor}
\usepackage{soul}
\usepackage{lipsum,microtype}
\usepackage{xcolor}
\usepackage{colortbl}
\usepackage[pagebackref,breaklinks,colorlinks]{hyperref}

\usepackage[capitalize]{cleveref}
\crefname{section}{Sec.}{Secs.}
\Crefname{section}{Section}{Sections}
\Crefname{table}{Table}{Tables}
\crefname{table}{Tab.}{Tabs.}

\def\paperID{10371} 
\def\confName{ICCV}
\def\confYear{2025}

\begin{document}

\title{ VIP: Video Inpainting Pipeline for Real World Human Removal  }

\author{Huiming Sun,Yikang Li, Kangning Yang, Ruineng Li, Daitao Xing, Yangbo Xie,\\ Lan Fu, Kaiyu Zhang, Ming Chen, Jiaming Ding, Jiang Geng, Jie Cai, Zibo Meng, Chiuman Ho }

\twocolumn[{
\renewcommand\twocolumn[1][]{#1}
\maketitle
\begin{center}
    \centering

    \includegraphics[width=1\textwidth]{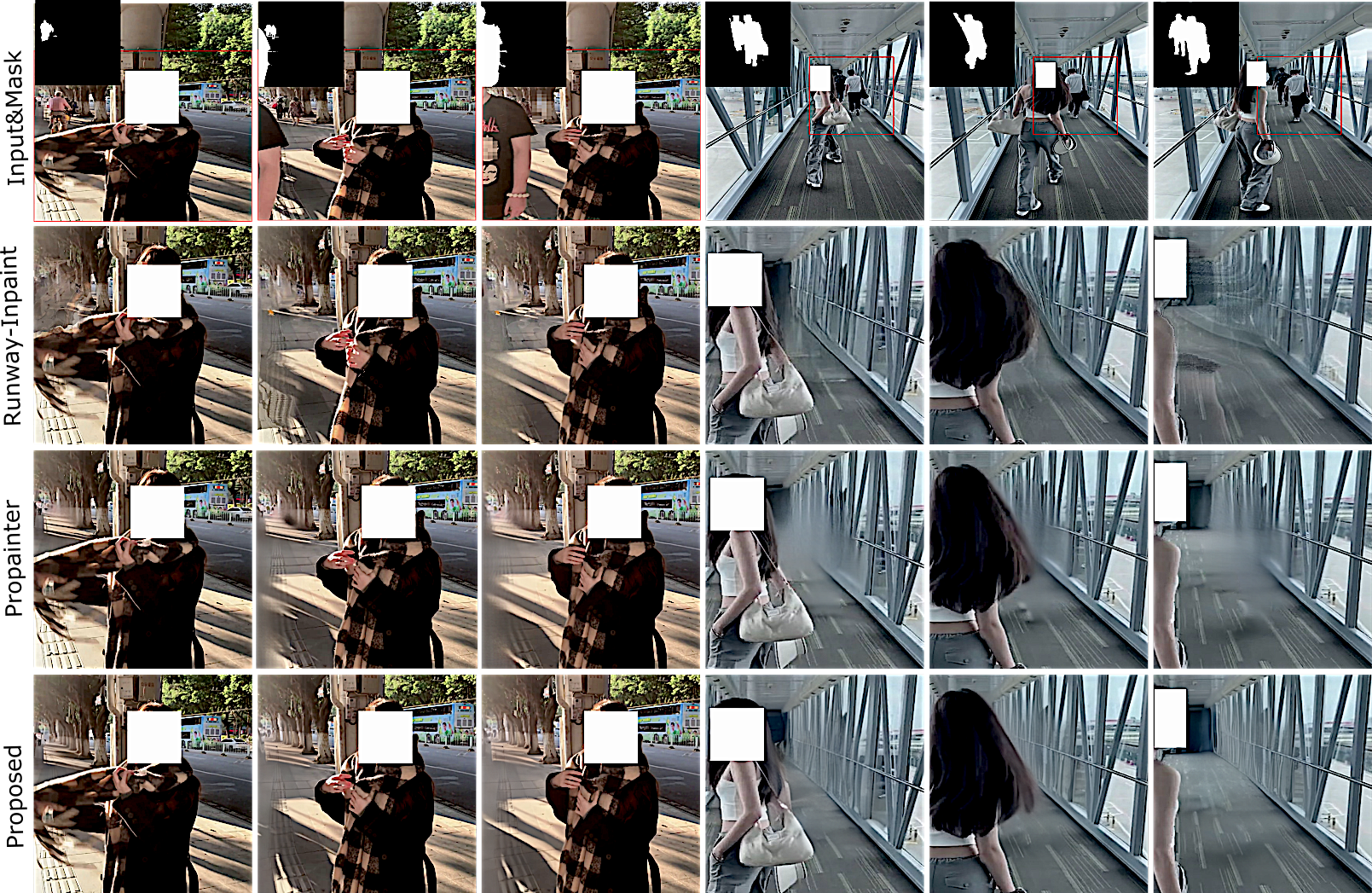}
    \captionof{figure}{
   Video inpainting result generated by VIP, the comparison showcases show its ability to generate better inpainting result.
    }
    \label{fig: Figure1}
\end{center}
}
]


\newcommand{\hm}[1]{\textcolor{red}{\hl{#1}}}
\begin{abstract}
Inpainting for real-world human and pedestrian removal in high-resolution video clips presents significant challenges, particularly in achieving high-quality outcomes, ensuring temporal consistency, and managing complex object interactions that involve humans, their belongings, and their shadows. In this paper, we introduce VIP (Video Inpainting Pipeline), a novel promptless video inpainting framework for real-world human removal applications. VIP enhances a state-of-the-art text-to-video model with a motion module and employs a Variational Autoencoder (VAE) for progressive denoising in the latent space. Additionally, we implement an efficient human-and-belongings segmentation for precise mask generation. Sufficient experimental results demonstrate that VIP achieves superior temporal consistency and visual fidelity across diverse real-world scenarios, surpassing state-of-the-art methods on challenging datasets. Our key contributions include the development of the VIP pipeline, a reference frame integration technique, and the Dual-Fusion Latent Segment Refinement method, all of which address the complexities of inpainting in long, high-resolution video sequences.
\end{abstract}

\section{Introduction}
\label{sec:intro}
Video inpainting, the task of reconstructing missing or undesired content in video sequences while maintaining spatio-temporal coherence, has garnered significant attention in the computer vision community due to its wide range of applications, such as object removal, video restoration, and film post-production~\cite{kim2019deep, wang2019video, liu2021fuseformer}. 
Despite the progress made, existing methods still struggle to achieve high-quality results while maintaining temporal consistency and handling complex object interactions within real-world high-resolution video contexts.

In this paper, we present VIP (Video Inpainting Pipeline), a novel video inpainting framework designed for real-world human removal in high-resolution videos without any prompt guidance. 
Building upon the state-of-the-art T2V (text-to-video) model~\cite{guo2023animatediff}, we apply a motion module to achieve high-quality, high-resolution video inpainting. Our approach utilizes a Variational Autoencoder (VAE) to encode both the input video and the masked video into a latent space, where progressive denoising is implemented by using spatial layers and novel motion modules to capture dynamic information and ensure temporal consistency. Additionally, an efficient human-and-belongings segmentation module is applied which can accurately identifies and segments human subjects along with their belongings and shadows, providing precise masks for high-resolution video inpainting. Our pipeline redefines conventional approaches by incorporating humans, their belongings, and their shadows as a cohesive instance for detection and segmentation.

Sufficient experiments in this paper demonstrate that VIP consistently outperforms current state-of-the-art methods, achieving superior temporal consistency and visual quality across a range of scenarios, particularly in real-world high-resolution videos. We evaluate our approach on the challenging YouTube-VOS-test dataset \cite{xu2018youtube} and a self-collected dataset of real-world videos (approximately 3 seconds each), showcasing its effectiveness in handling complex object interactions, dynamic motion, and crowded scenes. We claim four main contributions summarized as follows:
\begin{itemize}
\item We propose VIP, a noval video inpainting pipeline featuring an efficient segmentation and inpainting model, achieving high-quality human removal in real-world high-resolution videos without relying on text descriptions.
\item We introduce reference image integration with the inpainting inference process to substantially enhance the temporal consistency and smoothness.
\item A dual-fusion latent segment refinement is proposed in the inference stage to generate consistent long video contents.
\item Extensive experiments and user studies are conducted to validate the superiority of the proposed VIP, demonstrating its effectiveness in preserving spatio-temporal coherence and generating visually pleasing results.
\end{itemize}

\section{Related Work}

\textbf{Video segmentation} involves the process of partitioning a video sequence into multiple segments or objects to identify and track different entities or regions throughout the video~\cite{zhou2022survey}. Compared to segmentation in static images, video segmentation not only requires segmenting objects in individual frames but also needs to consider temporal correspondence and consistency across multiple frames. Numerous approaches have been proposed in recent literature via supervised~\cite{chen2020state, hu2017maskrnn, perazzi2017learning}, unsupervised, or semi-supervised learning paradigms. For instance, the MaskRNN~\cite{hu2017maskrnn} 
Li er al.~\cite{li2013video} propose an unsupervised
recent works on video segmentation exploit visual/text prompts in a video as reference to identify and segment target objects. 

\textbf{Video inpainting} is a crucial technique in computer vision, aimed at reconstructing missing or incomplete content in video sequences while maintaining spatial and temporal coherence~\cite{kim2019deep}. Traditional video inpainting methods often rely on patch-based approaches~\cite{huang2016temporally, newson2014video, wexler2007space}, which are often computationally expensive, have difficulties with non-repetitive content, and lack semantic understanding for complex scenarios. With the rise of deep learning, 3D convolution-based~\cite{chang2019free, chang2019learnable,sun2023coarse, hu2020proposal, wang2019video} and attention-based approaches~\cite{li2020short, oh2019onion, zeng2020learning, lee2019copy, liu2021fuseformer} provide more plausible and efficient solutions. 
For example, Chang et al.~\cite{chang2019free, chang2019learnable} first propose a learnable gated temporal shift module, and further extend it to combine 3D gated convolution with Temporal PatchGAN for video inpainting tasks. To better model long-range correspondences in video sequences, Zeng et al.~\cite{zeng2020learning} adopt an attention mechanism to search for coherent content from frames along the spatial and temporal dimensions, and introduce a joint spatial-temporal transformer network to fill in missing regions in video sequences. In addition, 
Some works~\cite{gao2020flow, xu2019deep} propose focusing on optical flow-based methods to exploit spatiotemporal consistency in videos. For example, FGVC~\cite{gao2020flow} first computes the completed flow fields, and then propagates video content across motion boundaries. Imagen Video~\cite{ho2022imagen} leverages a cascaded diffusion model to condition video generation on text descriptions. AVID~\cite{zhang2024avid} uses an image diffusion model and further designs a motion module and structure guidance to achieve fixed-length video inpainting. On this basis, it facilitates the inpainting with arbitrary length via a temporal MultiDiffusion sampling inference pipeline. Similar to AVID~\cite{zhang2024avid}, CoCoCo~\cite{zi2024cococo} additionally introduces a motion capture module to improve motion consistency.

\begin{figure*}[htbp]
\centering
\footnotesize
\includegraphics[width=1\textwidth]{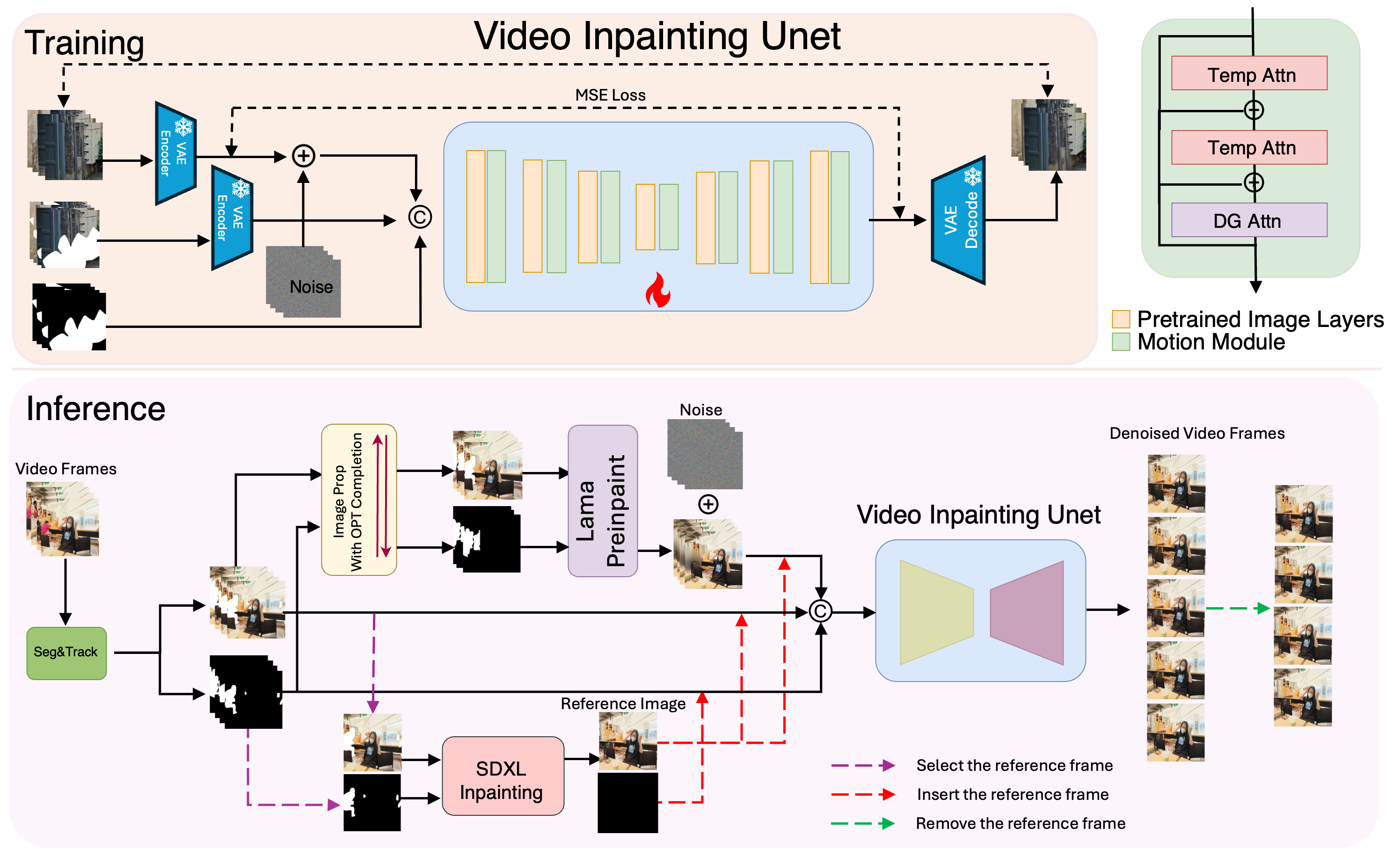}

\caption{The figure illustrates the training and inference processes for a Video Inpainting Unet model.  In the training stage, we employ 3 parts as Input: Latent, Mask and Mask Latent.  The bottom section depicts the inference pipeline. The inference process incorporates multiple stages of frame processing, including optical flow warpping and alignment, reference frame inpainting, and iterative inpainting steps before the final video inpainting unet. (For the sake of brevity, we omit the VAE encoding and deocde process in the inference pipeline.)}
\label{fig:pipeline}	 
\end{figure*}
\section{Methodology}

For a promptless video inpainting method, given a video sequence $X = \{X_{f} \in \mathbb{R}^{H \times W \times 3}\}^{F}_{f=1}$, our aim is to perform inpainting without using prompts and eliminate objects in the masked areas. We have divided this challenge into two main components:
1). High-quality mask generation: $M = \{M_{f} \in \mathbb{R}^{H \times W \times 1}\}^{F}_{f=1}$
2). Mask area denoising to produce high-quality inpainted video. Our method, which we call VIP (Video Inpainting Pipeline), is illustrated in Fig.~\ref{fig:pipeline} for the diffusion component. For the diffusion part, we build upon the T2V (text-to-video) model~\cite{guo2023animatediff}, enhancing it with a motion module for achieving high-quality, promptless video inpainting. For masks generation along the temporal axis, we implement the video object tracking algorithm \cite{cutie} with our segmentation module to generate mask sequences $\{M_0, M_1, M_2, ..., M_F\}$. The segmentation module is only utilized to generated masks in anchor frames $M_{anchor}$ where $N(M_{anchor}) << N(M_f)$, the $N(M_{anchor})$ is the number of anchor frames. All the remaining masks are generated by the propagation of the anchor masks with the video object tracking algorithm \cite{cutie}.


\subsection{Human Detection and Segmentation}
\label{sec:det-seg}
In this paper, the term ``human detection and segmentation" extends beyond traditional definitions to include the detection and segmentation of humans, their belongings, and their shadows as a unified instance. The paper provides basic information on our proposed human detection and segmentation approach; for a comprehensive overview of the pipeline, please refer to the supplemental materials.

\begin{figure}
\centering
\footnotesize
\includegraphics[width=1.0\linewidth]{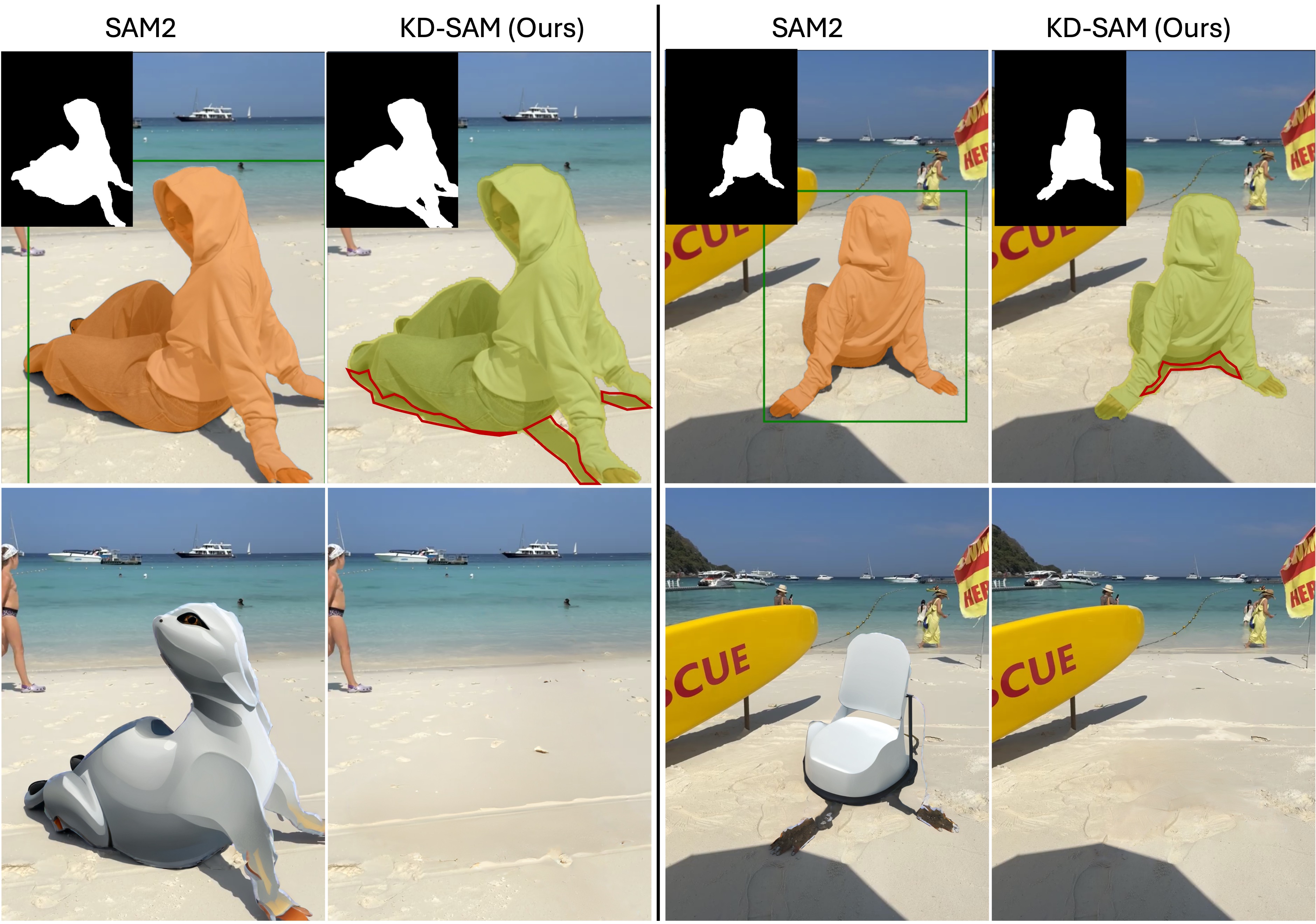}

\caption{Demonstration of our shadow detection and segmentation method) in comparison to the SAM2 image segmentation model). The red contours show the associated shadows detected and segmented by our algorithm. General-purpose segmentation models like SAM2 fail to accurately segment the corresponding shadows for the humans and lead to a bad generation. In contrast, our shadow detection and segmentation approach successfully segments and aligns shadows with the associated objects or humans, providing more precise and context-aware segmentation results which lead a perfect inpainting effect by SDXL inpainting model.}
\label{fig:kdsam-shadow}	 
\end{figure}

\noindent\textbf{Human Detection:}
\label{sec:huamn-det}
We select YOLOv9 \cite{yolov9} architecture as our detection module since the YOLO series is renowned for its efficiency and high accuracy in comparison to larger detection models. 
For human shadow detection, we integrate the shadow detection algorithm from \cite{silt} into our human detection pipeline. Some modifications are made to adopt the model architecture into more real-world cases. 
We also propose a human-shadow pairing strategy based on two key assumptions. 1). shadow masks associated with humans must not be excessively large or small; 2). shadow masks must exhibit a connection or overlap with the lower portion of the corresponding human figure. Some shadow detection and segmentation examples are demonstrated in Fig. \ref{fig:kdsam-shadow}.


\noindent\textbf{Human Segmentation:}
\label{sec:human-seg}
For the segmentation task, we employ the Segment Anything Model (SAM) \cite{sam} with only bounding box prompt as the foundational framework. To improve the model’s efficiency, we incorporate knowledge distillation techniques, as mentioned in \cite{efficientsam,mobilesam}, utilizing TinyViT \cite{tinyvit} as a compact vision encoder which is named as Knowledge Distilled SAM (KD-SAM). Additionally, we integrate a deformable attention module \cite{deformableattn} and high-quality token embedding from \cite{hqsam} into KD-SAM to enhance the ability of capturing small objects. 

\noindent\textbf{Human Tracking:}
We adopt the Cutie~\cite{cutie} algorithm as our tracking module, as its model architecture allows for seamless integration of the segmentation module within the whole tracking pipeline.


\subsection{The Overall Framework of Video Diffusion Inpainting}

\subsubsection{Training Stage}
During training, which is demonstrated in the upper row of Fig. \ref{fig:pipeline}, we utilize three input components similar to image diffusion networks:
1. Noise video clip $X^{1:F}$, 2. Mask video clip $M^{1:F}$, 3. Masked video clip $X^{1:F} \odot M^{1:F}$.

We employ a Variational Autoencoder (VAE) to encode both $X^{1:F}$ and $X^{1:F} \odot M^{1:F}$ into latent space, where we perform progressive denoising. Our model adapts spatial layers from previous work~\cite{Rombach_2022_CVPR}  and incorporates motion modules to capture dynamic information and ensure temporal consistency. To achieve promptless video inpainting, we fine-tune the entire U-Net architecture using only the generic ``inpainting'' text prompt during both training and inference. 
For the motion module architecture, we adopt the motion module proposed by CoCoCo~\cite{zi2024cococo}, while streamlining it by removing the cross-attention component. Our modified module comprises two temporal attention blocks and one damped global attention from ~\cite{zi2024cococo}. This architectural design enables more effective temporal consistency and significantly reduces spatial-temporal inconsistency in the generated sequences.

\noindent\textbf{Training Objectives:}
We train the Video Inpainting U-Net model in two stages. In the first stage, we use only the L1 loss on the latent codes. Given a video clip $\mathbf{x}^{1:f} \in \mathbb{R}^{f \times c \times w \times h}$ and its corresponding masked video clip $\mathbf{x'}^{1:f} = \mathbf{x}^{1:f} \odot \mathbf{M}^{1:f}$, they are encoded to latent codes $\mathbf{z}^{1:f}$ and $\mathbf{z'}^{1:f}$ frame-wise by a VAE encoder. The mask input $\mathbf{m}^{1:f}$ is resized to 1/8 scale to obtain the non-mask area, and we predict the added noise $\epsilon$. Our $\mathcal{L}_{r}$ loss for the latent codes is:
\begin{equation}
\begin{split}
    \mathcal{L}_{r} = w_1 \lVert \epsilon_\theta(\mathbf{z}^{1:f}, \mathbf{M}^{1:f}, \mathbf{z'}^{1:f}) - \mathbf{\epsilon}^{1:f} \rVert_1 \odot \mathbf{m}^{1:f} \\
    + w_2 \lVert \epsilon_\theta(\mathbf{z}^{1:f}, \mathbf{M}^{1:f}, \mathbf{z'}^{1:f}) - \mathbf{\epsilon}^{1:f} \rVert_1 \odot (1 - \mathbf{m}^{1:f})
\end{split}
\end{equation}
where $w_1$ and $w_2$ are the weighting factors for non-mask and mask areas, respectively, and $\epsilon_\theta$ represents the video inpainting U-Net function.
In the second stage, we use the post-VAE pixel reconstruction loss borrowed from~\cite{zhang2024pixel}. The latent codes are decoded back to pixel space by the VAE decoder $g$. Our pixel reconstruction loss is:
\begin{equation}
\mathcal{L}_{pixel} = \lVert \mathbf{x}^{1:f} - g(\mathbf{z_0}^{1:f}) \rVert_1
\end{equation}
The final training objective combines the latent $\mathcal{L}_{r}$ loss and pixel reconstruction loss:
\begin{equation}
\mathcal{L} = \mathcal{L}_{r} + \alpha \mathcal{L}_{pixel}
\end{equation}
where $\alpha$ is a hyperparameter balancing the two losses. We set $w_1 = 1$, $w_2 = 2$, and $\alpha = 3$ in our experiments.


\begin{figure}
\centering
\footnotesize
\includegraphics[width=1\linewidth]{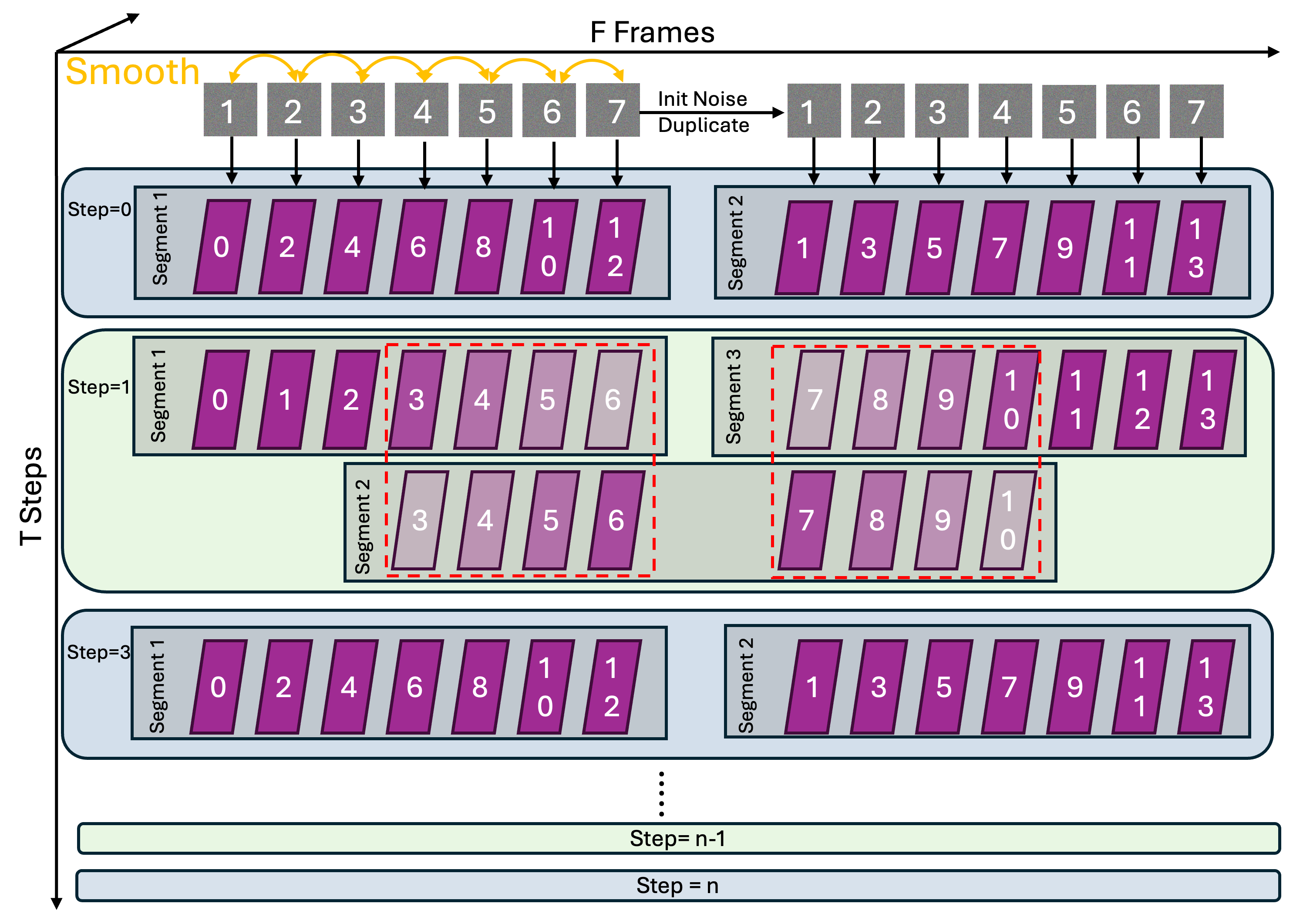}

\caption{ Dual-Fusion Latent Segment Refinement Visualization. }
\label{fig:longvideo}	 
\end{figure}

\begin{figure*}[htbp]
\centering
\footnotesize
\includegraphics[width=0.95\linewidth]{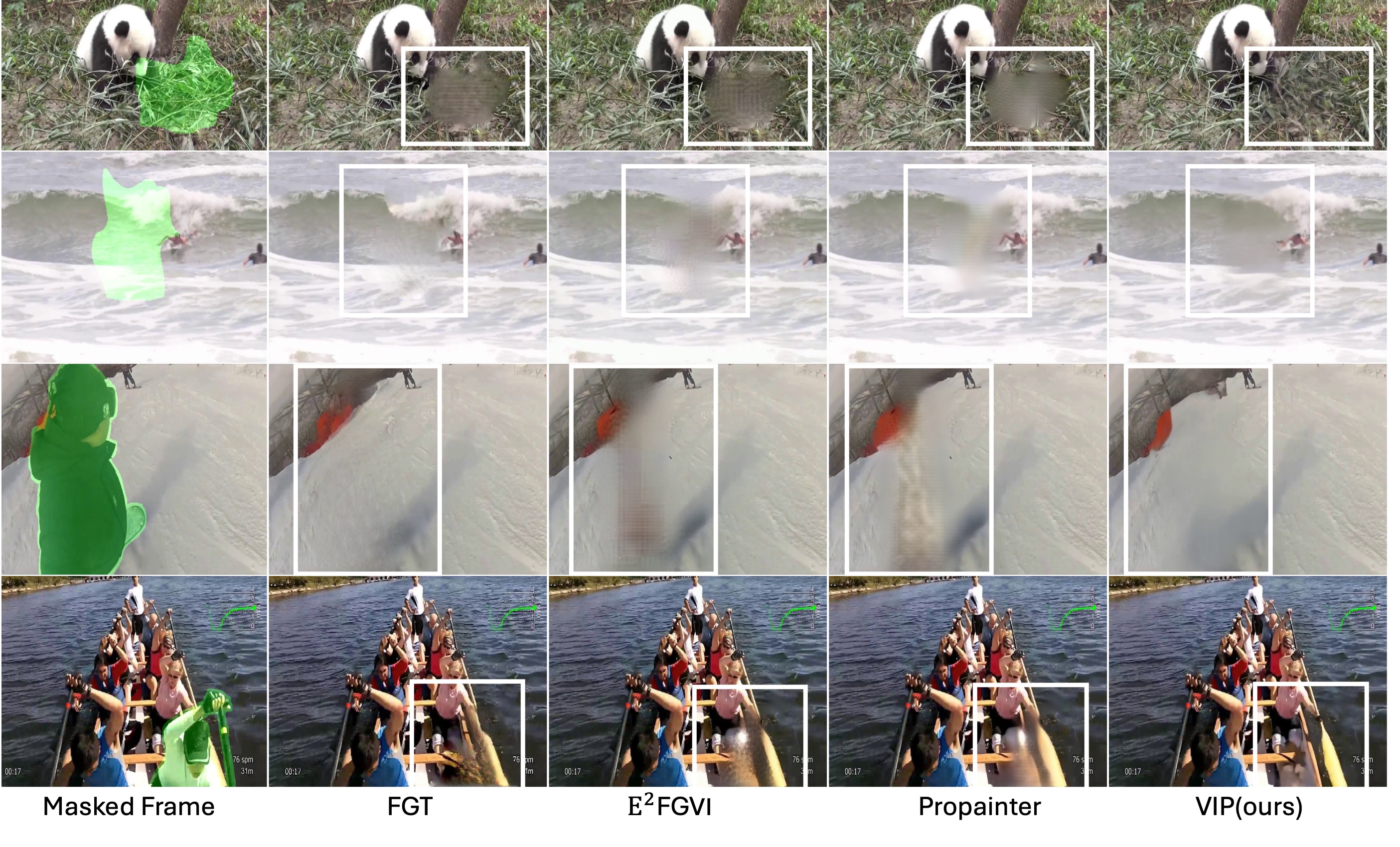}

\caption{  Qualitative comparisons on both video completion and object removal for high resolution videos.} 
\label{fig:qualitative}	 
\end{figure*}

\subsubsection{Inference Stage}
As shown in the lower row of the Fig \ref{fig:pipeline}, the inference stage mainly utilize two methods to enhance the performance of the high-resolution video inpainting.

\noindent\textbf{Optical Flow-Based Completion:}
Leveraging the characteristics of moving objects, we complete background information in certain scenes using optical flow. After obtaining $M^{1:F}$ through the KD-SAM pipeline for segmentation and tracking, we incorporate ProPainter's~\cite{zhou2023propainter} pixel propagation module to reduce the pressure of video inpainting and to maintain better temporal coherence to fill some missing areas by known information from adjacent frames. Most of small moving missing regions could be filtered out by this method. For regions that cannot be completed via optical flow, we employ the LAMA~\cite{suvorov2021resolution} image inpainting model   as a pre-inpainting method.

\noindent\textbf{Reference Image Integration:}
To address large mask missing areas in videos, we introduce a reference image approach. Given the effectiveness of existing image inpainting models like SDXL-inpainting~\cite{podell2023sdxl}  for complex scenarios, we select a reference frame $R_X$ and its corresponding mask $R_M$ from $X^{1:F}$ and $M^{1:F}$, respectively. We then apply SDXL inpainting to this frame. The inpainted reference frame is inserted into $X^{1:F}$ and $M^{1:F}$ before VAE encoding and removed before decoding. This approach helps propagate missing area information to other frames and enhances temporal smoothness, while avoiding direct frame replacement to maintain temporal continuity.

\begin{table*}
    \centering
    \footnotesize

    \caption{Quantitative comparison of different methods. Left: results on VOS-Test dataset. Right: results on Social Media dataset. The best and the second performance are marked in \textcolor{red}{red} and \textcolor{blue}{blue}. $E_{warp}^{*}$ denotes  $E_{warp}^{*} (\times 10^{-3}) $.   All methods are evaluated following their default settings, except we didn't resize the input video's size.}
    \begin{tabular}{l|ccccccccc||ccccc}
    \toprule
         &\multicolumn{9}{c||}{VOS-Test Dataset}& \multicolumn{5}{c}{Social Media Dataset}\\
         \cline{2-15}
         Methods&PSNR $\uparrow$& SSIM $\uparrow$& VFID $\downarrow$ & $E_{warp}^{*}$  $\downarrow$  & SC $\uparrow$ &BC $\uparrow$& TF $\uparrow$ &MS $\uparrow$  &CI $\uparrow$ & SC $\uparrow$ & BC$\uparrow$ & TF$\uparrow$ & MS$\uparrow$ &  CI$\uparrow$ \\ 
         \hline
        \multicolumn{15}{c}{\cellcolor{purple!10} Transformer-Based Inpainting Model} \\
        \hline
        FuseFormer~\cite{liu2021fuseformer} & 29.19 & 0.9328 & 0.068 & \textcolor{blue}{3.32}& 77.86 & 92.82 & 92.52 & 94.38 & 0.29 & 87.34 & 92.57 & 93.79 & 93.63  & 0.25\\
        ISVI~\cite{Zhang_2022_CVPR} & 31.41 & 0.9587 & 0.064 & 4.89& \textcolor{blue}{84.29} & 92.85 & 92.46 & 94.21 & 0.43 & \textcolor{blue}{88.83} & 93.01 & 93.14 & 93.17  & \textcolor{blue}{0.31}\\
        FGT~\cite{zhang2022flow} & \textcolor{blue}{33.15} & \textcolor{blue}{0.9669} & \textcolor{blue}{0.053} & 7.48& 81.68 & \textcolor{red}{92.90} & \textcolor{blue}{92.94} & \textcolor{blue}{94.71} & 0.29 & 88.03 & 92.77 & 93.61 & 93.45  & 0.26\\
        E$^{2}$FGVI~\cite{ liCvpr22vInpainting} & 30.83 & 0.9534 & 0.069 & 7.79& 79.40 & 91.93 & 92.13 & 94.56 & 0.34 & 87.83 & 92.30 & 93.79 & 93.63  & 0.30\\
        
        Propainter~\cite{zhou2023propainter} & \textcolor{red}{33.71} & \textcolor{red}{0.9681} & 0.056 & 10.85& \textcolor{red}{84.29} & \textcolor{blue}{92.85} & 92.46 & 94.21 & \textcolor{blue}{0.43} & \textcolor{red}{89.10} & \textcolor{red}{93.54} & \textcolor{blue}{93.87} & \textcolor{blue}{93.63} & 0.26\\
     \hline
\multicolumn{15}{c}{\cellcolor{purple!10} Diffusion-Based Inpainting Model} \\
\hline
        CoCoCo~\cite{zi2024cococo} & 28.24 & 0.9422&0.073&3.54&80.96&91.61&92.37&94.07&0.37&86.61&91.68&93.14&92.61&0.16\\
        VIP (ours) & 31.54 & 0.9578& \textcolor{red}{0.051} & \textcolor{red}{3.27}& 80.35 & 92.77 & \textcolor{red}{92.99} & \textcolor{red}{94.72} & \textcolor{red}{0.50} & 88.42 & \textcolor{blue}{93.26} & \textcolor{red}{93.93} & \textcolor{red}{93.64} & \textcolor{red}{0.50}\\
    \bottomrule
    \end{tabular}
    \label{tab:comparison}
\end{table*}

\subsubsection{Dual-Fusion Latent Segment Refinement For Long Video Generation  }
Computational constraints of video diffusion models pose significant challenges when handling such extended frame sequences. To address this limitation, recent approaches like MultiDiffusion~\cite{bar2023multidiffusion} and MimicMotion~\cite{mimicmotion2024} have introduced innovative latent fusion techniques. Yet, these methods are primarily tailored for video generation tasks and may not be directly applicable to the nuanced requirements of object removal. Unlike conventional Text-to-Video (T2V) or Video-to-Video (V2V) models that are limited to generating short sequences, real-world video inpainting tasks often involve processing longer durations, typically 3-4 seconds at 24 fps, resulting in approximately 72 frames. However, Diffusion Video inpainting presents a unique challenge where the majority of the frame content is known, except for the regions containing objects targeted for removal. The temporal dynamics of object appearances and disappearances further complicate this task, as targets may be present only in specific frames rather than consistently throughout the sequence. This scenario offers an opportunity to leverage background information from frames where the target is absent to reconstruct occluded areas in frames where it appears.

To address this problem, We propose ``Dual-Fusion Latent Segment Refinement" that leverages frame-wise noise patterns to enhance temporal coherence and computational efficiency. As shown in Fig.~\ref{fig:longvideo},  our method begins by initializing F frames' noise with a smooth noise progression, where each frame's noise is derived from its adjacent frames. This initialization ensures that neighboring frames share similar noise characteristics, promoting consistent denoising trajectories. The process is then duplicated with a slight offset to further reinforce temporal stability. Our diffusion process operates in T steps, with each step refining the frame representations. Notably, we introduce a segment-part-based processing technique that allows for parallel computation of frame subsets, significantly reducing the number of required diffusion passes. This approach can be flexibly extended to process every n-th frame simultaneously, balancing efficiency and temporal consistency. 


\section{Experiments}

The dataset and training detail we use to train the Video Inpainting Pipeline will be explained below.  \textit{Due to space limitations, human detection and segmentation are not our core contributions, what we want are those precise masks, which will be illustrated in the supplement material.}

\subsection{Datasets}

For the self-collected dataset targeting real-life scenarios, we gather ``4K city walk'' related videos, including city street walking, countryside walking, and shopping mall walking scenes, totaling 2.4M seconds. We then crop these into 0.24M clips, each 10 seconds long, and resize them to 1080p resolution. Additionally, we utilize the WebVid-10M~\cite{Bain21} and ACAV-100M~\cite{lee2021acav100m} datasets, filtering for high-resolution videos (larger than $512 \times 512$ pixels). We also use the image dataset LAION-5B~\cite{schuhmann2022laion} for image-video joint training. Please refer the supplemental materials for more detailed training information.

For the high-resolution evaluation set, we use the YouTube-VOS-test dataset~\cite{xu2018youtube}, including 547 videos ($1280\times720$). Furthermore, we self-collect 100 live photos, such as selfie videos ($720\times960$), each approximately 3 seconds long and containing 72 to 110 frames. We sample them all into 20 frames per video into training samples.

\subsection{Implementation Details}
We use Stable Diffusion 1.5-inpainting~\cite{Rombach_2022_CVPR} as the base text-to-image model to initialize our video diffusion model. We set the denoisied sequence length T as 24 and we apply random generate masks and existing segmentation masks as the mask input. We employ DDIM sampler, v-prediction strategy and AdamW optimizer to optimize the whole model. All the image and video training samples, we do 80\% random crop and 20\% resize to the target size.  During the inference, we set the denoising step number as 8 and didn't apply the classifier-free guidance. We train our model on $6 \times 8$  Nvidia A100 (80G) for around 1M steps. The total parameter for our model is 1.35B. The inference time for 24 frames is around 18 seconds on A100(80G) GPU.

\begin{figure}[h]
\centering
\footnotesize
\includegraphics[width=1\linewidth]{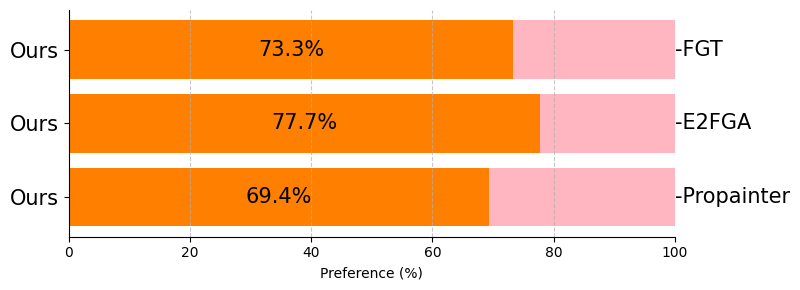}

\caption{ User prefers VIP over other methods. } 
\label{fig:user_study}	 
\end{figure}

\subsection{Evaluation and Comparison}
Traditional image quality metrics like PSNR and SSIM are inadequate for evaluating object removal tasks due to the absence of ground truth reference images in masked regions. Moreover, these pixel-wise metrics may penalize perceptually plausible results that deviate from the original content. The reference metrics like PSNR and SSIM can not evaluate the quality of object removal tasks without paired Input and GT.  To address these limitations, we use a dual-track evaluation framework  that assesses both temporal coherence and frame-level quality. For temporal assessment, we adopt VBench~\cite{huang2023vbench} metrics including Subject Consistency (SC), Background Consistency (BC), and Temporal Flickering (TF). For frame-level evaluation, we leverage Co-Instruct~\cite{wu2024openended} to perform win-rate analysis between methods.


\begin{table}
    \centering
    \footnotesize
    \caption{Ablation study of inference pipeline module. $E_{\text{warp}}^{*}$ denotes $E_{\text{warp}}^{*} (\times 10^{-3})$.  OP means Optical Flow-Based Completion, R means Reference Image Integration.}
    \begin{tabular}{lcc|cccc}
        \toprule
        Model& OP &  R & PSNR $\uparrow$ & SSIM $\uparrow$ & VFID $\downarrow$ & $E_{\text{warp}}^{*}$ $\downarrow$ \\
        \midrule
        VIP& &  & 30.72 & 0.9511 & 0.052 & 3.35 \\
        VIP&\checkmark & & 30.19 & 0.9488 & 0.056 & 3.40 \\
        VIP&& \checkmark & 31.19 & 0.9566 & 0.055 & 3.27 \\
        VIP&\checkmark& \checkmark& \textbf{31.54} & \textbf{0.9578} & \textbf{0.051} & \textbf{3.27} \\
        \bottomrule
    \end{tabular}
\label{tab:ablation_study}
\end{table}

\paragraph{Quantitative Evaluation:} We compare our VIP method with 6 state-of-the-art methods. They are FuseFormer~\cite{liu2021fuseformer}, ISVI~\cite{Zhang_2022_CVPR}, FGT~\cite{zhang2022flow}, E$^2$FGVI~\cite{liCvpr22vInpainting}, Propainter~\cite{zhou2023propainter} and also diffusion-based inpainting model CoCoCo~\cite{zi2024cococo} which set the input prompt as ``no human" for inpainting area. As shown in Table~\ref{tab:comparison}, our method achieves competitive performance across multiple metrics. Specifically, VIP demonstrates strong temporal consistency with the highest TF score of 92.99 and competitive BC score of 92.77. For frame-level assessment, our method achieves the best CI score of 0.50, indicating superior perceptual quality in the inpainted regions.  Our method shows better performance in motion-smoothness metrics (MS: 94.72) and temporal stability measures. This suggests that VIP effectively balances both spatial fidelity and temporal coherence, particularly in handling dynamic scenes and complex object removals.

\paragraph{Qualitative Evaluation:} Fig.~\ref{fig:qualitative} shows visual comparisons between our method and previous approaches on various challenging scenarios. Compared to existing methods, VIP demonstrates superior performance in preserving both spatial details and temporal consistency. While previous methods may generate visible artifacts or temporal flickering in complex scenes, our approach produces more natural and coherent results, especially in challenging cases involving dynamic motion, complex textures, and crowded scenes. The visual results align with our quantitative findings, particularly in terms of temporal stability and perceptual quality.

\paragraph{User Study:}
To validate our quantitative and qualitative results, we conducted a comprehensive user study evaluating the perceptual quality of our inpainting results. We randomly sampled 25 test cases from the VOS-test dataset and 25 from our social media dataset for evaluation. The study compared our method against state-of-the-art approaches including FGT~\cite{zhang2022flow}, E$^{2}$FGVI~\cite{liCvpr22vInpainting}, and ProPainter~\cite{zhou2023propainter}. For each test case, we presented users with three versions of the same video: the ground truth, our result, and the result from one competing method, with the order of our method and the competitor randomized to eliminate bias. Ten participants were asked to select their preferred result between the two inpainted versions. As shown in Fig.~\ref{fig:user_study}, our method achieved a preference rate of 70\%--78\%, demonstrating the superiority of our VIP inpainting approach and validating the effectiveness of our proposed evaluation metric.
\subsection{Ablation Study}

\begin{figure}[htbp]
\centering
\footnotesize
\includegraphics[width=0.95\linewidth]{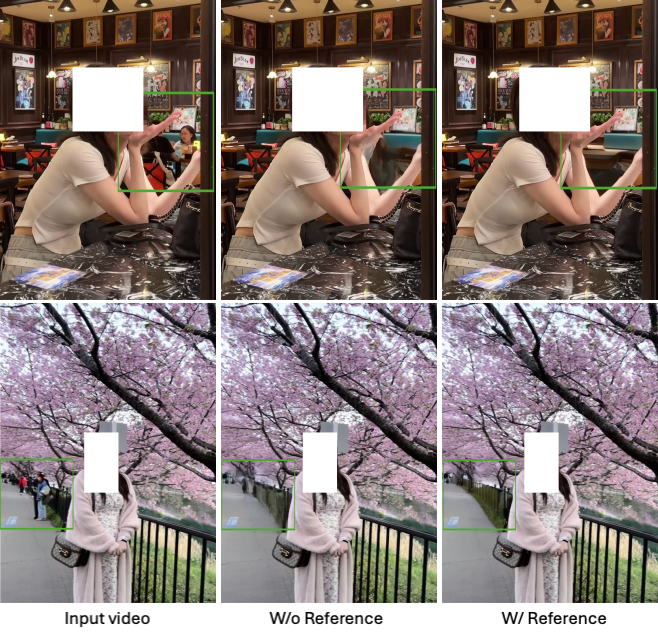}

\caption{  Comparison of w/ and w/o image reference. Zoom in for more details in the images. } 
\label{fig:reference}	 
\end{figure}

\begin{figure*}[htbp]
\centering
\footnotesize
\includegraphics[width=0.95\linewidth]{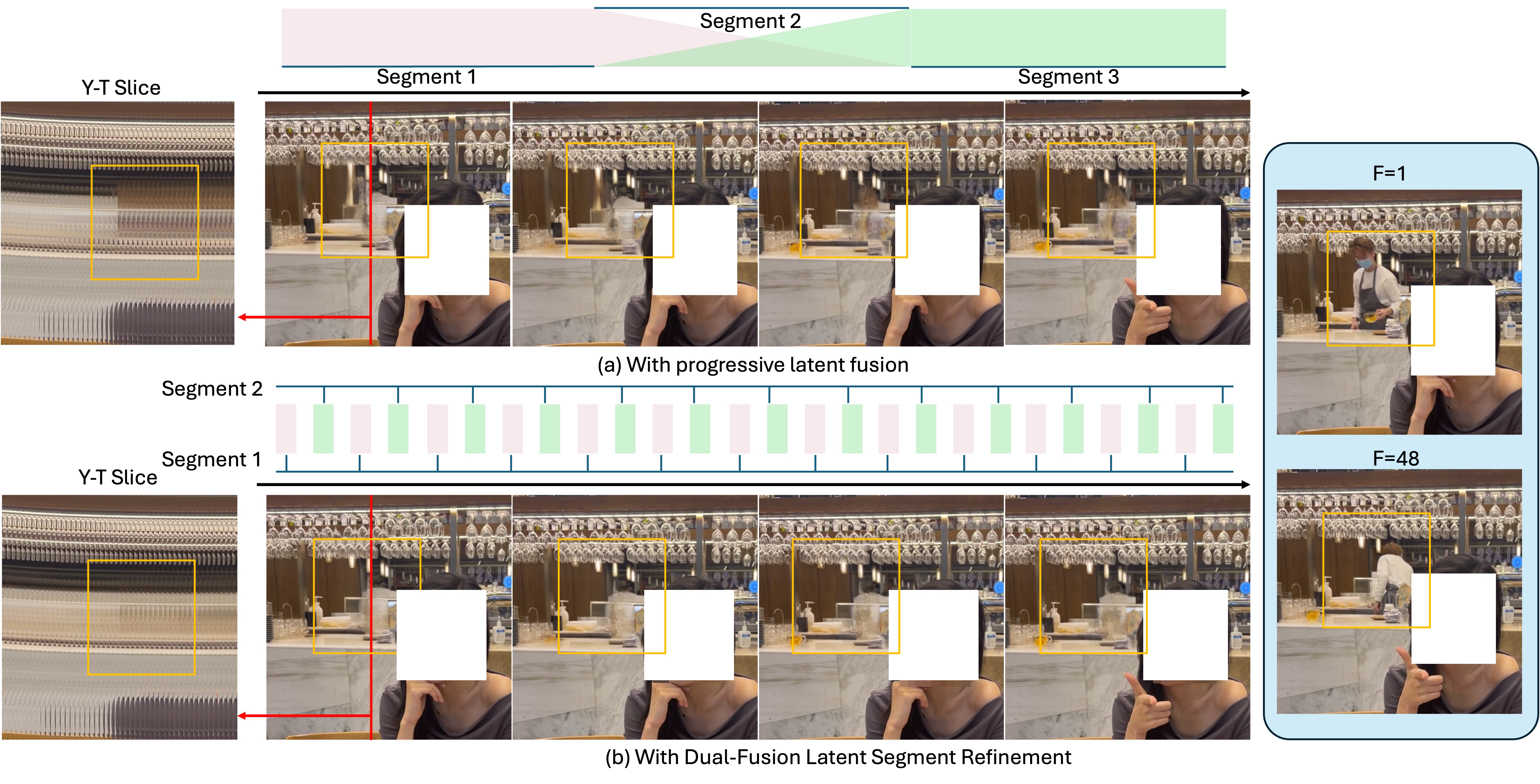}

\caption{ Dual-Fusion latent segment refinement transitions for long video inpainting (48 frames). (a) The vertical strips in the Y-T slice figure shows progressive latent fusion causes temporal discontinuity. (b) Dual-Fusion latent segment refinement transitions lead soomth transitions. Zoom in for more details in the frames.  } 
\label{fig:longvideo_compare}	 
\end{figure*}

\noindent\textbf{Inference Pipeline Components:} Our two key inference pipeline modules: Optical Flow-Based Completion (OP) and Reference Image Integration (R). As shown in Table.~\ref{tab:ablation_study}, both modules contribute positively to the overall performance. The baseline model without either module achieves a PSNR of 30.72 and SSIM of 0.9511. Adding only Reference Image Integration slightly decreases performance, likely due to the challenge of maintaining temporal consistency when using single-frame guidance. In contrast, using only Optical Flow-Based Completion shows notable improvements , indicating its effectiveness in preserving temporal coherence. The combination of both modules achieves the best overall performance (PSNR: 31.54, SSIM: 0.9578) while maintaining competitive warping error ($E_{warp}^{*}$: 3.27 $1\times10^{-3}$). This suggests that the two modules complement each other effectively, with OP providing temporal consistency and R enhancing spatial detail quality.

\noindent\textbf{Reference Frame:} Fig.~\ref{fig:reference} demonstrates the effectiveness of reference frame guidance in our approach. While our model uses only 1.3B parameters and is not specifically optimized for image inpainting, we achieve improved performance on challenging cases with large occlusions by leveraging SDXL-inpainting capabilities. As shown in the two examples, without reference frames, the inpainting results can be either inconsistent or semantically reasonable but visually suboptimal. By incorporating reference frame guidance, our video inpainting method successfully propagates well-reconstructed regions across the temporal dimension.

\noindent \textbf{Analysis of Dual-Fusion Latent Segment Refinement:} For long-duration video inpainting tasks (i.e., object removal), diffusion-based video models face a critical challenge in maintaining temporal consistency. Unlike video generation tasks, video inpainting benefits from strong prior knowledge of the surrounding context, enabling a more efficient generation process with fewer diffusion steps compared to pure Gaussian noise initialization. However, this efficiency introduces a new challenge: while the generated content may be visually plausible, even slight temporal mismatches can be perceptually jarring to human observers.

We also observe a phenomenon: imperfect generations that maintain precise alignment with the masked regions often appear more visually coherent than higher-quality generations with minor temporal discontinuities. Based on this observation, we propose the Dual-Fusion Latent Segment Refinement method, illustrated in Fig.~\ref{fig:longvideo_compare}. Our approach employs the same video inpainting model but introduces a novel fusion strategy that prioritizes temporal coherence by maximizing the temporal extent of segments while enforcing harmony constraints.

As shown in the Y-T slice visualization (left), the baseline only progressive latent fusion approach~\cite{mimicmotion2024} exhibits sudden object changes that result in visible artifacts and temporal discontinuities. In contrast, our method achieves smoother transitions and superior visual quality, as demonstrated in the central frames. Furthermore, our approach is computationally efficient, requiring progressive latent fusion only at steps 1 and 7 within an 8-step sequence, resulting in a 75\% reduction in the number of fusion operations compared to the baseline method.
The qualitative results in Fig.~\ref{fig:longvideo_compare} demonstrate that our Dual-Fusion approach successfully addresses both temporal consistency and computational efficiency, producing more visually pleasing results.

\section{Conclusion}
In this paper, we presented VIP, a noval promptless video inpainting framework for real-world high-resolution human removal applications, introducing several key innovations: a reference frame integration technique that enhances inpainting quality, and  Dual-Fusion Latent Segment Refinement method that enables temporally consistent inpainting for longer video sequences. Through extensive experiments, our approach achieves superior performance in temporal consistency and visual quality across diverse scenarios without relying on text prompts, representing a significant progress in real-world product-level video inpainting applications.

{\small
\bibliographystyle{ieee_fullname}
\bibliography{egbib}
}

\end{document}